\documentclass[letterpaper, 10 pt, conference]{ieeeconf}  

\IEEEoverridecommandlockouts                              

\usepackage{cite}
\usepackage{graphicx}    
\usepackage{booktabs}    
\usepackage{array}        
\usepackage{multirow}     
\usepackage{amsmath}     
\usepackage[section]{placeins}      
\usepackage{hyperref}
\usepackage{cleveref} 
\usepackage{amssymb}
\providecommand{\BIBentryALTinterwordspacing}{}
\providecommand{\BIBentrySTDinterwordspacing}{}

\title{\LARGE \bf
SSR-ZSON: Zero-Shot Object Navigation via Spatial-Semantic Relations within a Hierarchical Exploration Framework
}

\author{Xiangyi Meng, Delun Li, Zihao Mao, Yi Yang and Wenjie Song*
\thanks{*This work was partly supported by Program for National Natural Science Foundation  of  China  (Grant  No.  62373052),  Beijing  Natural  Science Foundation (Grant No. 4252051), and in part by the National Key Laboratory of Science and Technology on Space Born Intelligent Information Processing TJ-01-22-09.}%
\thanks{The authors are with the School of Automation, Beijing Institute of Technology, Beijing 100081, China, (Corresponding author: Wenjie Song(email: songwj@bit.edu.cn)}%
}

\begin{document}

\maketitle
\thispagestyle{empty}
\pagestyle{empty}

\begin{abstract}

Zero-shot object navigation in unknown environments presents significant challenges, mainly due to two key limitations: insufficient semantic guidance leads to inefficient exploration, while limited spatial memory resulting from environmental structure causes entrapment in local regions. To address these issues, we propose SSR-ZSON, a spatial-semantic relative zero-shot object navigation method based on the TARE\cite{2021TARE} hierarchical exploration framework, integrating a viewpoint generation strategy balancing spatial coverage and semantic density with an LLM-based global guidance mechanism. The performance improvement of the proposed method is due to two key innovations. First, the viewpoint generation strategy prioritizes areas of high semantic density within traversable sub-regions to maximize spatial coverage and minimize invalid exploration. Second, coupled with an LLM-based global guidance mechanism, it assesses semantic associations to direct navigation toward high-value spaces, preventing local entrapment and ensuring efficient exploration. Deployed on hybrid Habitat-Gazebo simulations and physical platforms, SSR-ZSON achieves real-time operation and superior performance. On Matterport3D\cite{Matterport3D2017} and Habitat-Matterport3D\cite{HM3D2021} datasets, it improves the Success Rate(SR) by 18.5\% and 11.2\%, and the Success weighted by Path Length(SPL) by 0.181 and 0.140, respectively, over state-of-the-art methods.

\end{abstract}

\section{INTRODUCTION}

Robotic object navigation algorithms, as one of the core capabilities in industrial, search-and-rescue, and service scenarios, require robots to efficiently search for targets in unknown environments lacking prior maps. Although object navigation technologies have advanced rapidly, existing research still struggles to balance semantic reasoning and spatial exploration efficiency in open-world settings. Traditional geometry-driven exploration strategies often overlook semantic information, failing to achieve true goal-oriented navigation. Classical object navigation algorithms, meanwhile, rely excessively on environmental semantics while neglecting the profound influence of scene structures on exploration. This leads to insufficient adaptability in generalized scenarios and low task success rate.
    
Recent research in zero-shot object navigation has integrated LLMs into navigation systems to enhance semantic understanding. These approaches typically involve parsing textual instructions to generate relevant target information or modeling object associations via knowledge graphs to guide robot navigation. However, these methods face two primary challenges: First, they often neglect the impact of scene structure on exploration. While LLMs offer strong open-vocabulary reasoning capabilities, these are not effectively translated into actionable exploration decisions in unknown spaces, leading to a disconnect between semantic associations and path planning. Second, the heuristic value of semantic distributions in guiding target search remains underutilized. Current strategies primarily rely on object or scene correlations and traditional exploration planning methods, such as boundary or sampling point generation, without adequately integrating semantic insights with exploration strategies, resulting in suboptimal navigation efficiency.

\begin{figure*}[htbp] 
\centering
\includegraphics[width=\linewidth]{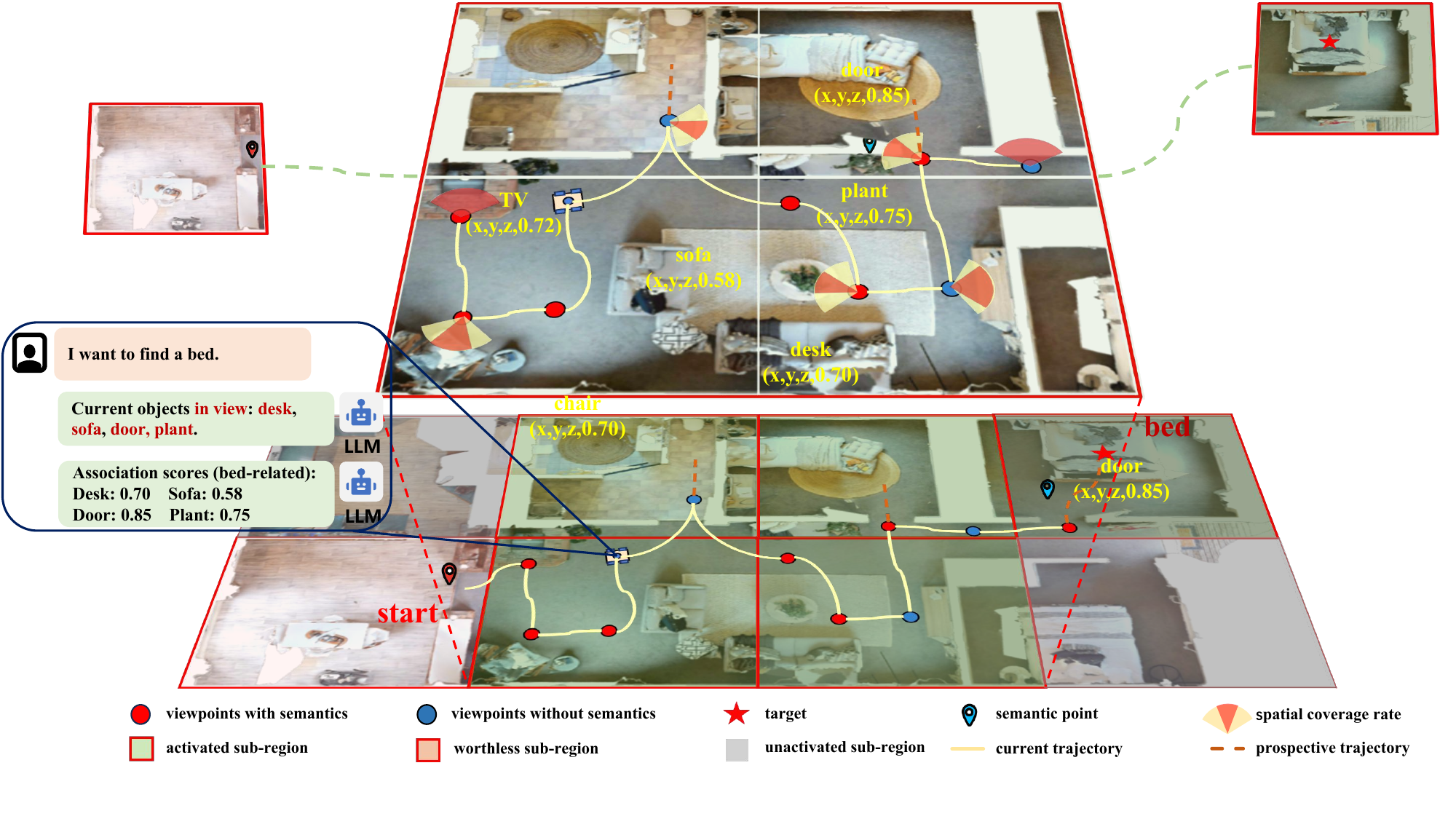} 
\caption{\textbf{Spatial-Semantic Collaborative Exploration Framework.} This framework integrates semantic reasoning based on LLMs with spatial exploration. A dynamic semantic topological map assesses the relevance of objects to the goal in real-time, while the environment is discretized into grid regions. Within each selected region, candidate viewpoints are generated based on local semantic density and coverage. These regions are then prioritized for exploration according to the number of viewpoints and their semantic relevance, with priority given to those exceeding a predefined threshold. The exploration path is formed by connecting traversable candidate viewpoints. This framework drives robotic navigation towards regions with high semantic value and geometrically unknown areas.}
\label{figure1}
\end{figure*}
    
To address these challenges, this paper introduces the SSR-ZSON framework, as shown in \cref{figure1}. First, the natural language processing capabilities of LLMs are leveraged to analyze human instructions and identify the target object to be located. Simultaneously, by constructing a dynamic semantic topology map, LLMs are utilized to evaluate in real-time the association weights between visible objects and the target objects, such as reasoning the ``co-occurrence probability of a monitor and a keyboard'' or the ``likelihood of a bed appearing in a bedroom'' and so on. Second, the environment is discretized into uniform grid sub-regions, and a regional priority scoring function is designed. This function calculates the unexplored surface coverage rate and semantic association value of each sub-region to determine an optimal exploration sequence, enabling bidirectional exploration of high-semantic-value and geometrically unknown areas. In addition, multiple viewpoints are evenly distributed within each selected region, and a two-stage viewpoint screening mechanism is implemented. Initially, a candidate viewpoint set is generated based on semantic density and uncovered surface coverage; subsequently, connectable viewpoints are sampled from this set and linked to form an exploration path for the region.

In summary, our key contributions are as follows.

1) \textbf{Viewpoint Generation Strategy:} We design a viewpoint generation strategy that optimizes both spatial coverage and semantic density, guiding the robot toward areas of semantic relevance but underexplored to significantly reduce ineffective exploration.

2) \textbf{Dynamic Semantic Weighting-Based Region Exploration:} We introduce a hierarchical exploration approach that partitions the environment into several regions, employs LLMs to dynamically assign semantic weights and structural priors, and incorporates region-wise state tracking to enhance long-term exploration memory.

3) \textbf{Viewpoint-Region Joint Object Navigation Framework:} We propose a viewpoint-region integrated zero-shot object navigation framework that tightly couples semantic inference with spatial-semantic decision-making, effectively mitigating the spatial-semantic decoupling issue in unknown environments.

\section{Related Work}

\subsection{Spatial Perception-based Autonomous Exploration}%
Spatial perception-based autonomous exploration relies on the perception of environmental geometric features. Classical algorithms can be categorized into three types: frontier-based, sampling-based, and hybrid approaches. Frontier-based exploration, first proposed by B. Yamauchi\cite{Yamauchi1997}, identifies boundaries between known and unknown regions as navigation targets and iteratively expands these boundaries to complete exploration. However, this distance-prioritized strategy often leads to local optima. Subsequent improvements incorporate dynamic replanning and multi-criteria evaluation functions. In 2017, Umari proposed the Rapidly-exploring Random Tree (RRT) method\cite{1998Rapidly} to detect frontiers\cite{Umari2017}. When a tree edge straddles both unknown and known regions, the point possessing a path from the unknown region to the known region constitutes the frontier. In complex large-scale scenarios, numerous subsequent approaches have proposed combining frontier-based and sampling-based strategies by structuring the overall planning process into global and local hierarchies. For instance, to ensure efficient exploration while preventing entrapment in local minima, \cite{Selin2019EfficientAE} adopts the Receding Horizon Next-Best-View planning(RH-NBVP) method\cite{Bircher2016} as the local planner alongside a frontier exploration-based method       as the global planner. The TARE method\cite{2021TARE} employs viewpoint sampling to jointly optimize path length and unknown space coverage across global and local planning layers. The HPHS method\cite{HPHS2024} integrates a hybrid frontier sampling approach with hierarchical planning strategies, reducing the complexity of planning while enhancing the efficiency of autonomous exploration. Exploration based on spatial perception is sensitive to geometric structure, but it cannot efficiently navigate for specific target task requirements.

\subsection{Semantic Information-driven Object Navigation}%
Object navigation typically utilizes object detection for semantic annotation and trains navigation policies through Reinforcement Learning(RL). Zhu et al. first conceived the concept of a target-driven model\cite{Zhu2017}, which serves as the predecessor to object navigation. This approach was developed to tackle challenges concerning goal and scene generalizability while enabling straightforward implementation of end-to-end navigation. Santhosh Kumar Ramakrishnan\cite{PONI2022} et al. proposed a novel network architecture that predicts dual complementary potential functions from semantic maps, leveraging supervised learning to train the network for target-driven navigation. \cite{GoT2024} employs a Goal-guided Transformer-based deep reinforcement learning strategy, significantly enhancing data efficiency during learning. Knowledge graph technology structures commonsense knowledge (e.g., object functionalities) to support task-level reasoning. The HOGN-TVGN\cite{YANG2024102671} embeds prior knowledge into reinforcement learning policies. With the continuous progress of large model technology, the field of object navigation has gradually introduced the reasoning ability of LLMs or MLLMs for cost evaluation. At the same time, with the powerful natural language processing ability of large models, human-robot interaction can also be realized.
For instance, \cite{zhou2024navgpt,NavCoT} build a visual-language navigation model based on LLMs, enabling information exchange with open-world environments. Additionally, the maturity of large-scale model technology has also driven the development of various zero-shot object navigation algorithms. The ZSON method\cite{ZSON2022} pioneers zero-shot object navigation, requiring no labeled data or demonstrations, and completely eliminates dependency on closed vocabulary. \cite{CoWs2023,Dorbala2024,Wenzhe2024,long2024instructnav,VLFM2024,VoroNav2024,UniGoal2025} achieve zero-shot object navigation without requiring task-specific training, prebuilt maps, or environmental priors. The above methods are highly dependent on semantic information, but lack complex processing of spatial information, and are easy to fall into local difficulties.

As evidenced by substantial prior research, numerous semantic-based object navigation methods focus their core objectives at the visual-semantic level, while opting for simplified frontier-based exploration or basic path planning algorithms at the spatial-structural level to guide robots. For instance, \cite{corl2023,L3MVN2023} employ a simplified frontier-based exploration approach integrated with semantic information, while maintaining a 2D environmental map. For each frontier region, they leverage an LLM to evaluate and select the optimal frontier, significantly enhancing the efficiency of object navigation. Specifically, for each candidate frontier region, an LLM-generated relevance score is assigned based on semantic associations (e.g., object-context co-occurrence likelihood). The frontier with the highest composite score is selected as the next exploration target, thereby enhancing target-oriented exploration efficiency. In contrast, our method integrates the coverage of spatial surfaces with the distribution and relevance of open-world semantic information. It operates respectively at both the viewpoint-level and region-level hierarchies, thoroughly integrating spatial and semantic information. These components interlock closely, jointly guiding the robot to complete object navigation tasks under zero-shot conditions.

\section{METHOD}

Our method proposes a collaborative navigation framework integrating spatial cognition and semantic reasoning. The system leverages a point cloud map constructed and maintained in real-time by laser-based SLAM as its foundation. Through uniform partitioning of the environment, a spatial grid map is established. Simultaneously, LLMs-driven semantic priority evaluation model (Sec. 3.1) generates relevance scores for target objects, forming a dynamically updated semantic associative topological graph.The core design centers on the hierarchical viewpoint-region collaborative evaluation model (Sec. 3.2), which screens high-value observation poses using dual viewpoint-level metrics: geometric coverage and semantic density, and activates sub-region exploration sequences based on region-level semantic association strength and spatial distribution. Further, we introduce the coverage-Aware sub-regions memory model (Sec. 3.3) that optimizes sub-region traversal logic via a state-switching mechanism based on the camera's viewing coverage to mitigate redundant exploration. The overall framework establishes a closed-loop ``Perception-Reasoning-Decision-Memory'' process. The spatial grid map facilitates efficient boundary detection and region segmentation, while the semantic topological map provides global and local information gain. Final viewpoint generation, region activation, and dynamic memory attenuation are coordinated to prioritize exploration of high-semantic value regions, optimizing object navigation efficiency.

\subsection{Semantic Prioritisation Evaluation Based on Large Language Models}
This module proposes a target object relevance calculation method that integrates spatiotemporal feature analysis with semantic reasoning, aiming to enhance target navigation efficiency in complex environments. The system employs a three-layer processing architecture: first, fusing multi-frame perceptual data through a spatiotemporal buffering mechanism, then performing semantic relevance reasoning using LLMs, and finally establishing an "object-relevance" memory through dynamic cache optimization to improve real-time performance.

The spatiotemporal buffer module uses a sliding window algorithm to store the detection results, and the hash key value $k\ =\ \left(\left\lfloor\frac{x}{0.1}\right\rfloor\ ,\ \left\lfloor\frac{y}{0.1}\right\rfloor\right)$ to spatially discretize the semantic point data, this mechanism effectively solves the problem of dense item group detection interference, such as detecting multiple adjacent chairs in an office scenario, and avoids redundant reasoning about duplicate items by retaining the detection result with the highest confidence within each grid cell.

\begin{figure}[htbp]
\vspace{-0.1cm} 
\centering
\includegraphics[scale=0.35]{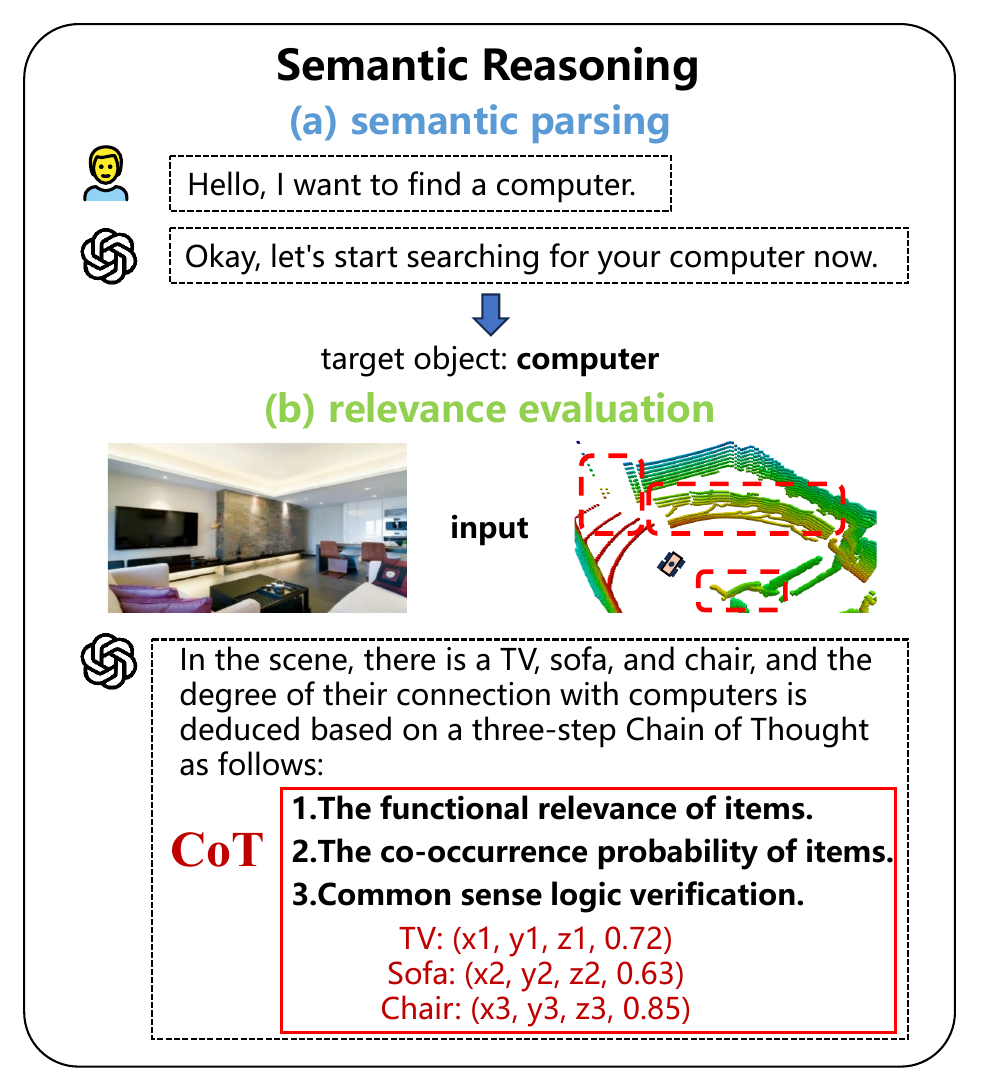}
\caption{\textbf{semantic reasoning level.} (a) Semantic parsing section: The LLM processes natural language to derive the target item.(b) Relevance evaluation section: Taking the current frame's visual field image and coordinates as input, this module evaluates the relevance of all detectable items within the view to the target item, along with their global coordinates.}
\label{figure2}
\vspace{-0.0cm} 
\end{figure}

As shown in \cref{figure2}, the semantic reasoning process integrates two key modules: semantic parsing and relevance evaluation, each handled by specialized models to enhance modularity and inference speed. Semantic parsing is crucial for human-robot interaction. Structured few-shot exemplars are designed to concretize natural human utterances into target object identifiers $e_{\text{target}}$ (e.g., parsing ``Help me find the fire\_extinguisher'' to ``fire\_extinguisher''). For relevance evaluation, a three-level chain-of-thinking evaluation model is constructed: firstly, scene localization is performed to establish a strong mapping relationship between the target items and a typical scene , then the scene topology analysis is performed to determine the topological connectivity strengths of all the detected items within the scene, and finally, the degree of functional coupling between items is verified, and based on the first two constraints, further scoring is performed according to the degree of functional further scoring based on the degree of association based on the first two constraints. The system adopts a dynamic scoring constraint mechanism to ensure that the scoring results conform to the laws of the physical world through a rigid hierarchical division. The above model is designed to guide the large model in output of standardized scores by designing structured cue word templates, and finally the temperature coefficient of the model is reduced to ensure the stability of the inference.

The system adopts a two-layer cache optimization strategy: in order to reduce the reasoning time brought about by the interaction step, the LRU cache is used. In order to reduce the reasoning time that may be brought about by the interaction step, a certain number of recent semantic parsing results are maintained through the LRU caching mechanism, and the timeliness is constrained to be 24 hours, which ensures that the parsing of the system on the language can be run for a long time. Dynamic deletion of records. The real-time processing module fuses the spatiotemporal buffer data with the semantic scores to generate point cloud data containing 3D coordinates and confidence:
\begin{equation}
P=\left\{\left(x_i,y_i,z_i,\max{\left(\alpha S_s, I_{\mathrm{target}}\right)}\right)\right\}
\end{equation}
where $S_s$ denotes the semantic relevance, $\alpha$ denotes the dynamic relevance parameter, and $I_{\mathrm{target}}$ is the minimum semantic relevance, which is guaranteed not to be zero.

\subsection{Hierarchical Viewpoint-Region Collaborative Evaluation Model}
This method constructs a two-tier assessment system at the viewpoint level and the regional level, forming a collaborative decision-making assessment model from the viewpoint level to the regional level.

As shown in \cref{figure3}: 
\begin{figure}[htbp]
\centering
\includegraphics[width=\linewidth]{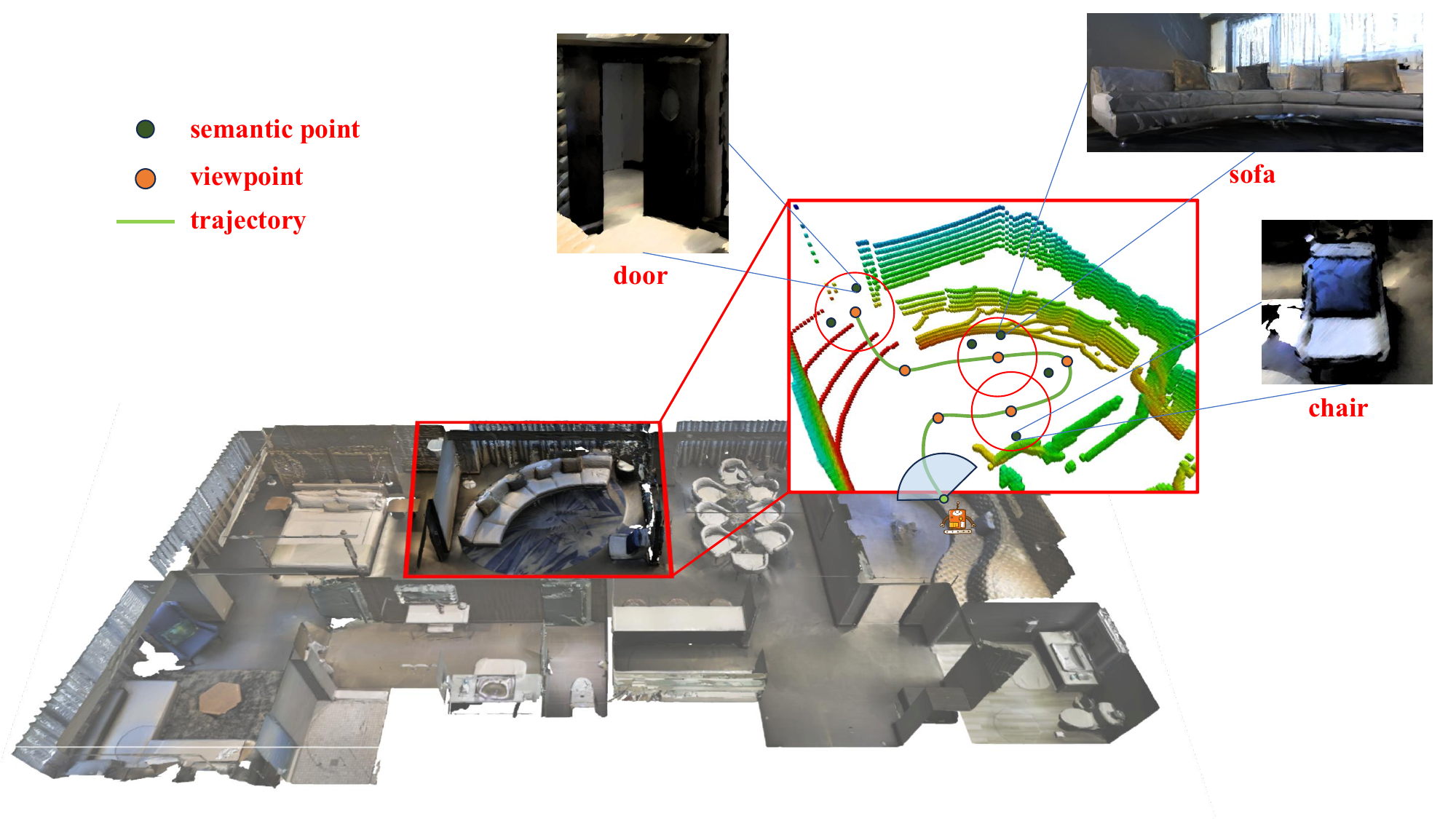} 
\caption{\textbf{viewpoint screening level.} Viewpoints (orange points) are randomly generated within the current accessible area. Each viewpoint undergoes dual screening based on geometric coverage and semantic density. Viewpoints encompassing higher concentrations of semantic points (dark green points) and newly captured uncovered surfaces (voxel point clouds scanned for the first time) within the red circles correlate with a higher probability of selection.}
\label{figure3}
\vspace{-0.3cm} 
\end{figure}
At the viewpoint screening level, a large number of viewpoints are randomly sampled in the accessible area, and the sampled viewpoints are screened by the dual indexes of geometric coverage and semantic density. Geometric coverage counts the amount of point clouds on uncovered surfaces from viewpoints under a certain perspective, and the higher the number, the higher the spatial exploration value,the expression is shown below:
\begin{equation}
S_{\mathrm{cov}}\left(v_j\right)=N_{\mathrm{unobserved}}\left(v_j\right)
\end{equation}
Semantic density counts the number of valid semantic points within a certain range for each viewpoint, the higher the density the higher the semantic exploration value,the expression is shown below:
\begin{equation}
S_{\mathrm{sem-density}}\left(v_j\right)=\sum_{p_k\in\mathcal{N}_r\left(v_j\right)} I\left(p_k\in P_{\mathrm{semantic}}\right)
\end{equation}
where $\mathcal{N}_r\left(v_j\right)$ denotes that within the cylindrical neighbourhood (height set to infinity, height not considered) of viewpoint $v_j$ with radius $r$, $P_{\mathrm{semantic}}$ is the set of all item points detected, and the $I\left(\bullet\right)$ function is used to count the number of effective number of items, and the meaningless detection words are eliminated by the large language model.
From the above, the evaluation function for each viewpoint is as follows:
\begin{equation}
S_{\mathrm{viewpoint}}\left(v_j\right)=\lambda_1\cdot S_{\mathrm{cov}}\left(v_j\right)+\lambda_2\cdot S_{\mathrm{sem-density}}\left(v_j\right)
\end{equation}
where $\lambda_1$ and $\lambda2$ are dynamic coefficients, adjusted based on spatial structures and semantic distributions across scenarios, to achieve a user-defined dynamic balance between geometric coverage and semantic density in the evaluation function.

\begin{figure}[htbp]
\vspace{-0.3cm} 
\centering
\includegraphics[width=\linewidth]{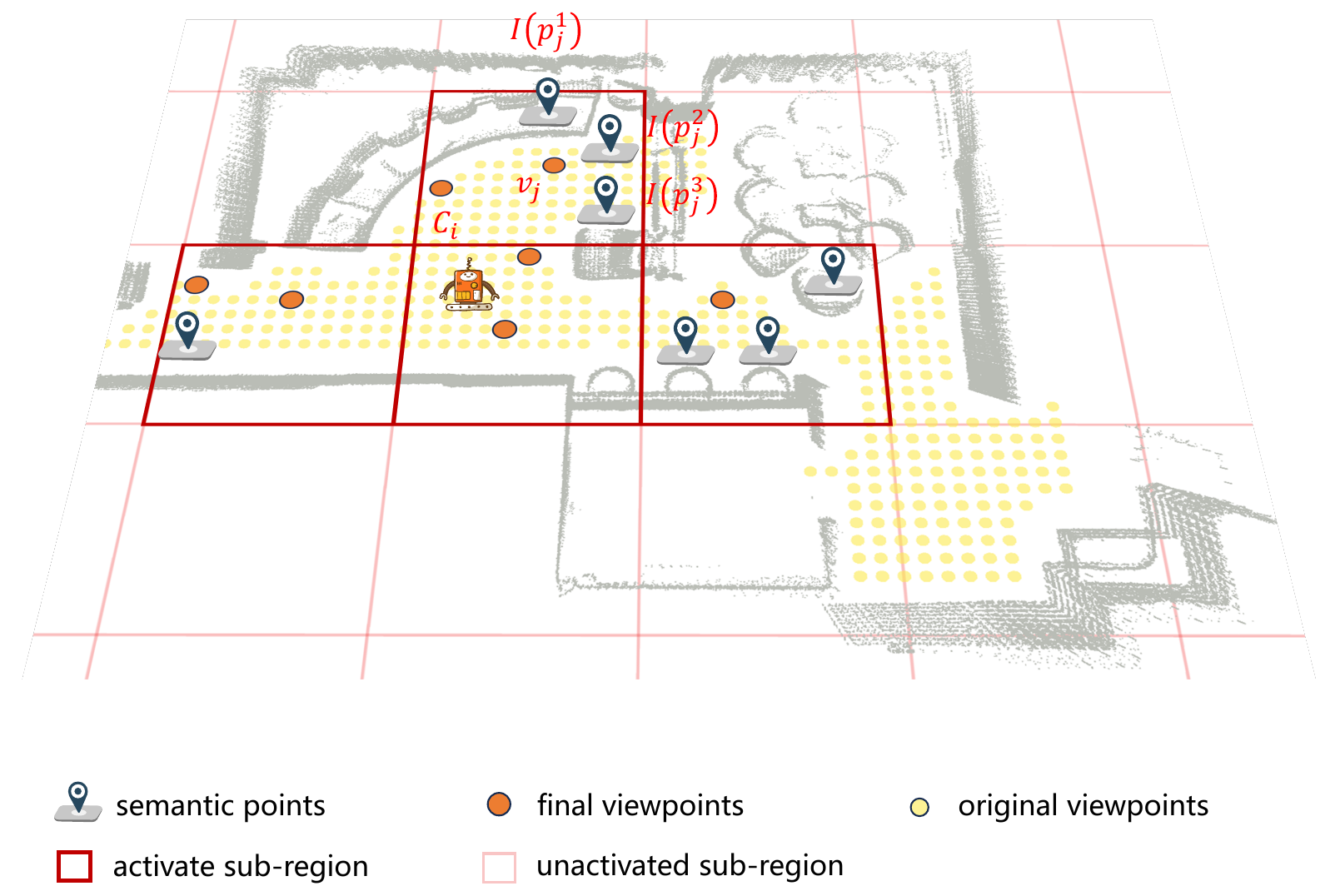} 
\caption{\textbf{region activation layer.} In sub-regions of predetermined size, a higher density of viewpoints correlates with closer proximity of semantic points within the viewpoints' vicinity and elevated relevance scores. This collective metric elevates the sub-region's overall score. Sub-regions persistently exceeding the scoring threshold are designated as activated, with the activated sub-region list undergoing dynamic updates.}
\label{figure}
\vspace{-0.3cm} 
\end{figure}

At the region activation layer, the sub-region scores are determined by the number of viewpoints within the sub-region as well as the semantic association scores. The sub-region semantic association score is weighted by scoring each viewpoint within a certain range of neighbourhood, and the total semantic association score is obtained by summing up the scores of all viewpoints in the sub-region, which is divided into two phases; the first phase searches globally with the maximum radius (the connecting circle in the sub-region), and locates the highly-associated targets based on the descending order of the semantic strength, and returns the results directly if there are points exceeding the threshold value; The second stage searches for semantic points within a certain range of neighbourhood of the viewpoint, and calculates the Gaussian decay weight of the viewpoint to get the semantic association score of the viewpoint, and we define $\omega_j^i$ as the weight value of the $i$-th semantic point in the neighbourhood of the viewpoint $v_j$, and then the formula of the weight function is shown as follows:
\begin{equation}
\omega_j^i = \exp\left(-\frac{\|p_j^i - v_j\|^2}{\sigma^2}\right), \quad \sigma = \frac{\sqrt{2}}{2} r_{xy}
\end{equation}
where $p_j^i=(p_{jx}^i,p_{jy}^i)$ are the coordinates of the semantic points and $v_j=(v_{jx},v_{jy})$ are the coordinates of the viewpoints,$r_{xy}$ is the radius of the neighbourhood of the viewpoint.

As a result, the semantic relevance of the $i$-th semantic point in the neighbourhood of viewpoint $v_j$ is defined as $I\left(p_j^i\right)$, and the formula of the semantic relevance evaluation function for each viewpoint is shown below:
\begin{equation}
{\bar{S}}_{v_j}=\frac{\sum_{i}{\omega_j^iI\left(p_j^i\right)}}{\sum_{i}\omega_j^i}
\end{equation}
As a result, the evaluation function formula for the ith sub-region $C_i$ is shown below:

\begin{equation}
{\bar{S}}_{C_i}=\left(\sum_{v_j\in C_i}{S_{\mathrm{viewpoint}}\left(v_j\right)}\right)+{\bar{S}}_{v_j}
\end{equation}

When ${\bar{S}}_{C_i}$ exceeds the threshold, the subregion is activated and it is included in the list to be explored. Upon acquiring the list of regions to be explored, in light of the total semantic weighted scores of the sub-regions within the list, a greedy strategy is employed. The traversal sequence of the sub-regions is computed in descending order of scores, thus deriving the regional-level guiding path. The sub-regions in the list of regions to be explored will be updated dynamically in accordance with the scores, constantly steering the robot towards regions featuring denser semantic information, higher semantic relevance weights, and larger unexplored surfaces.

\subsection{Coverage-Aware Sub-regions Memory Model}

For the activated sub-regions, we set the state switching function, adding the ``worthless'' state in the inactive and active states, and realize the memory management of the environment sub-regions through real-time perception and coverage evaluation based on Field of View (FOV) during the robot navigation. When the robot updated its pose each time, only the current sub-region was accumulated by discrete sampling according to the plane sector field of view of the camera. When the coverage ratio reaches the threshold when it is detected to leave the sub-region, it will be marked as ``worthless'' at the moment of leaving, so that it will not be selected in the subsequent global planning.

Each sub-region is uniformly subdivided into $N \times N$ grid cells. In every update of robot pose, we only process the current grid cell: using a planar sector FOV model with radius $R$ and angle $\alpha$, we mark grid cells that fall within the sector as ``seen''. This yields a monotonically increasing set of seen grid cells, used to approximate how much of the grid cell has actually been scanned. The coverage ratio is defined as:
\begin{equation}
    \mathrm{cov} = \frac{|\mathcal{S}_{\text{seen}}|}{N^2}
\end{equation}
where $\mathcal{S}_{\text{seen}}$ is the set of grid cells that have been marked as seen. A grid cell is considered seen by the current FOV if it lies within the planar sector (ignoring height):
\begin{equation}
    \mathrm{Visible}(g) = \mathbb{I}\Big[ \|\mathbf{v}_g\| \le R \ \wedge\ \angle(\mathbf{h}_t, \mathbf{v}_g) \le \alpha/2 \Big]
\end{equation}
where $\mathbf{v}_g$ is the 2D vector from the robot to the grid cell center and $\mathbf{h}_t$ is the 2D heading estimated from the last two robot positions (if the displacement is too small, the previous heading is reused). State switching is checked only at the moment the robot leaves a sub-region. If the accumulated coverage ratio of the just-left sub-region reaches the threshold $\tau$, that sub-region is set to ``worthless''; otherwise its state remains unchanged: If $\mathrm{cov}$ is greater than or equal to $\tau$, we switch the sub-region to the ``worthless'' state. Upon entering a new sub-region, we activate its state, reset the coverage bitmap to prevent interference between adjacent grids, and promptly perform a FOV accumulation step based on the current pose. The state transition based on view scanning can enhance the robot's memory during the target navigation process, prevent redundant exploration of the same areas, and thereby improve the overall efficiency of object navigation.

\section{Experimental Results}

The proposed method was validated in both simulation environments and real-world experiments. Through comparisons with multiple current state-of-the-art methods, the metrics utilized by our approach were superior to those of the current mainstream methods.
\subsection{Simulation Experiment}
\begin{figure}[htbp]
\centering

\begin{minipage}[b]{\linewidth}
    \centering
    \includegraphics[width=\linewidth]{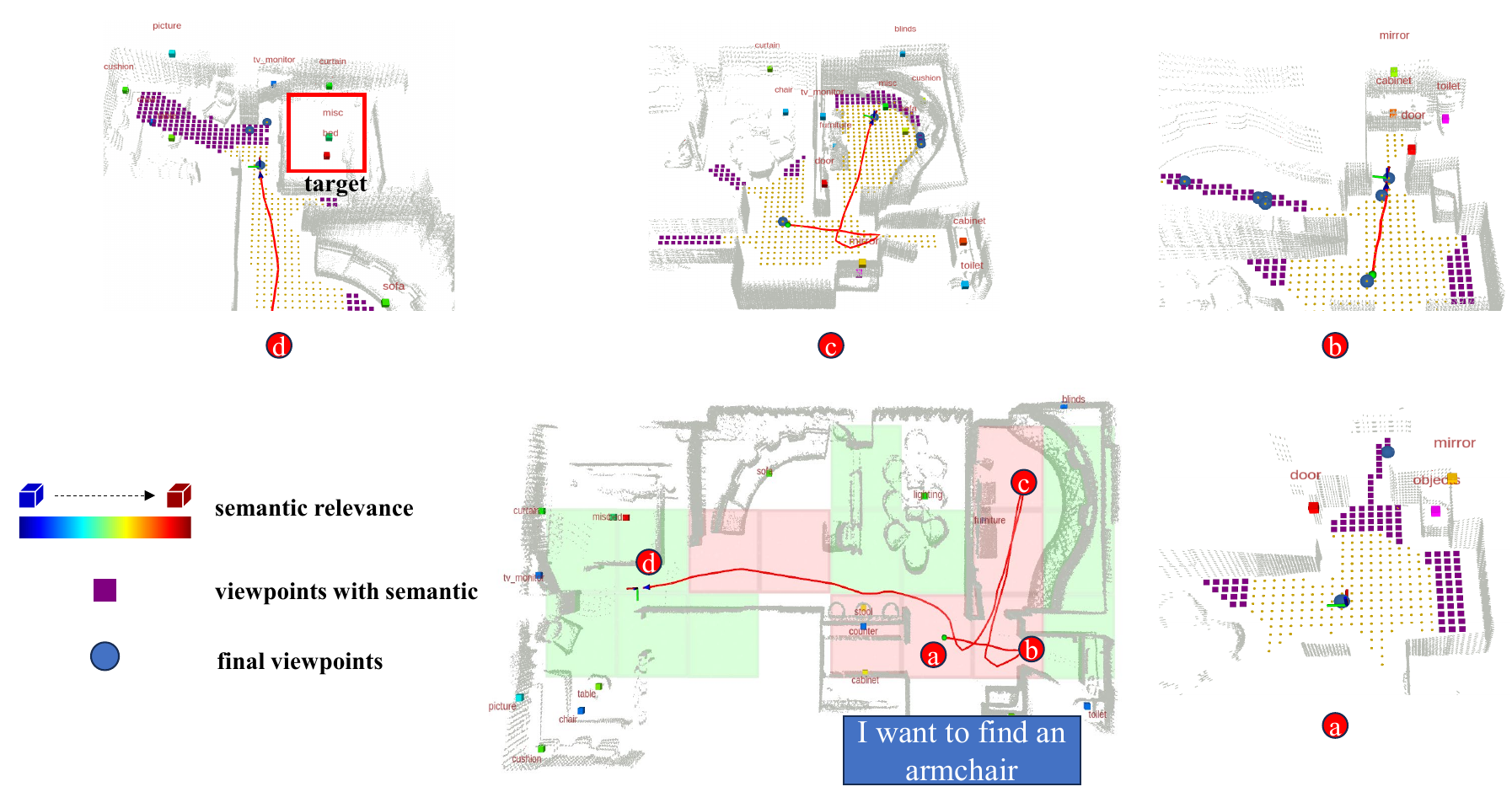} 
    \par\vspace{0.2cm} 
    \textbf{(a) MP3D} 
\end{minipage}

\vspace{0.8cm} 

\begin{minipage}[b]{\linewidth}
    \centering
    \includegraphics[width=\linewidth]{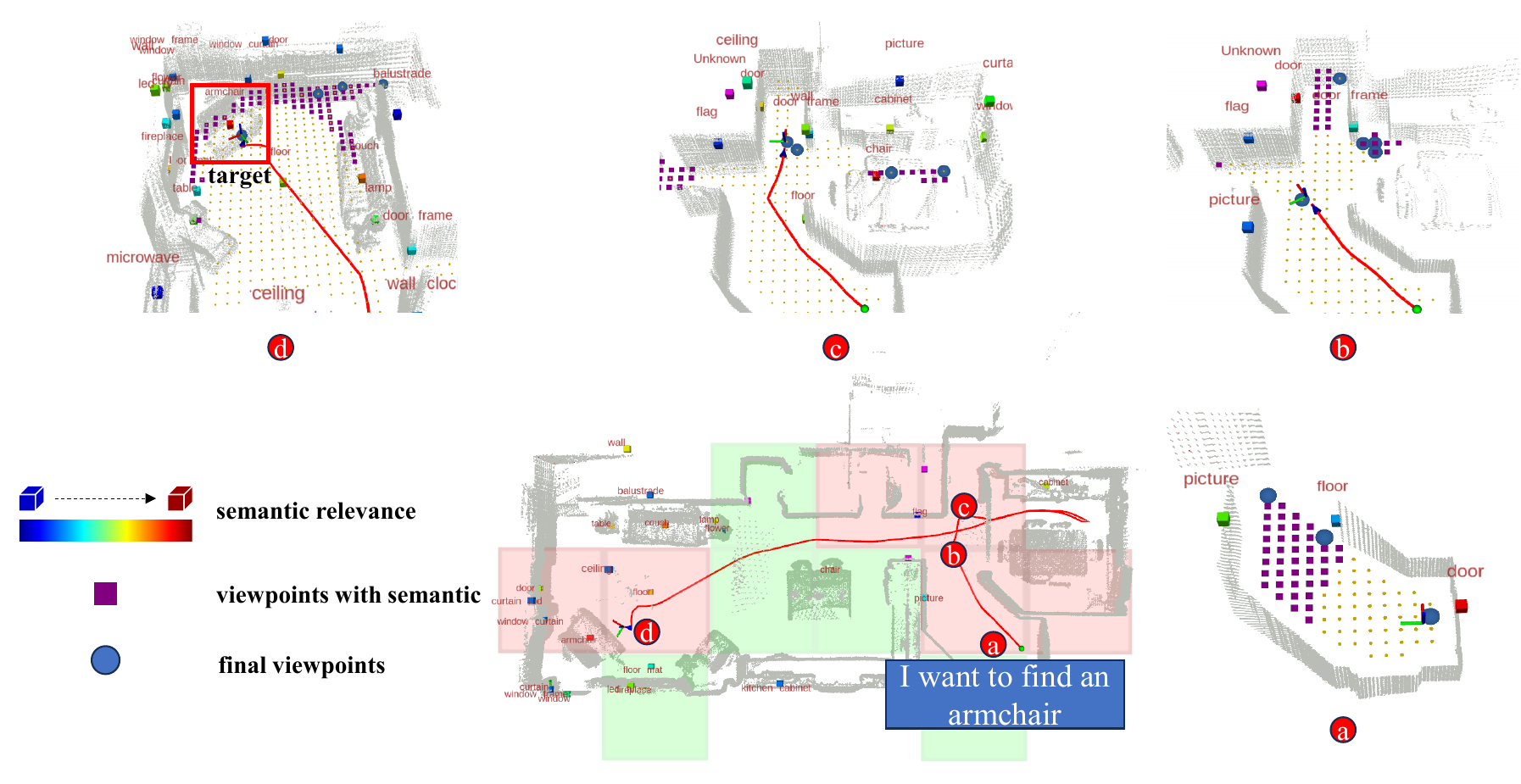} 
    \par\vspace{0.2cm} 
    \textbf{(b) HM3D}
\end{minipage}
\caption{\textbf{the outcomes of the simulation experiments.} In the figure, the yellow point clouds on the ground denote the original sampling viewpoints, the purple point clouds represent the sampling viewpoints under semantic weighting, and the blue spheres indicate the final viewpoints obtained from the current screening process. The path will be connected based on the final viewpoints and updated dynamically.}
\label{figure5}
\end{figure}

We carried out numerous tests within the combined simulation environment of Habitat and Gazebo, employing the MP3D\cite{Matterport3D2017} dataset and the HM3D\cite{HM3D2021} dataset. For the simulation experiments, semantic information and a simulated spatial environment are essential. Therefore, it is necessary to convert the semantic datasets into point clouds to acquire solid models. The semantic information is loaded in Habitat, while the spatial information is synchronously retrieved in Gazebo. The outcomes of the simulation experiments are presented in \cref{figure5}.

Our method was compared against the following five previous zero-shot goal navigation methods, evaluated using the metrics SR and SPL.The comparison methods included:

\textbf{ZSON}\cite{ZSON2022}: Proposes an open-world navigation novel paradigm that supports free-form linguistic instructions, enabling the accomplishment of target navigation tasks without requiring training.

\textbf{InstructNav}\cite{long2024instructnav}: A general instruction navigation system capable of executing different types of instruction in unknown environments.

\textbf{VLFM}\cite{VLFM2024}: Proposed a zero-shot method capable of performing goal-driven semantic navigation without task-specific training, pre-built maps, or environmental prior knowledge.

\textbf{VoroNav}\cite{VoroNav2024}: A zero-shot goal navigation framework based on Voronoi diagrams. 

\textbf{UniGoal}\cite{UniGoal2025}: A universal zero-shot goal-oriented navigation framework that unifies different goal types (including object categories, instance images, and text descriptions) via a unified graph representation method.

Considering the characteristics of the MP3D\cite{Matterport3D2017} and HM3D\cite{HM3D2021} datasets, more than five trials were carried out per scene, each targeting a different object object. That is, over five distinct target items were selected for experimentation per scene.The experimental results are shown in \cref{tab:vl_nav_performance}.

To test the system’s capability for natural language understanding, several different descriptive formats were selected. For instance, goal instructions such as ``Help me find the monitor'', ``I want to find a table'', and ``Someone might need a chair'' were used. This ensured that the system could correctly interpret the natural language commands and identify the target object.
\begin{table}[htbp]
\centering
\caption{Vision-language Navigation Performance (SR/SPL)}
\label{tab:vl_nav_performance}
\begin{tabular}{lcccc}
\toprule
\multirow{2}{*}{Method} & \multicolumn{2}{c}{SR (\%)} & \multicolumn{2}{c}{SPL} \\
\cmidrule(r){2-3} \cmidrule(l){4-5}
 & MP3D & HM3D & MP3D & HM3D \\
\midrule
ZSON & 15.3 & 25.5 & 0.048 & 0.126 \\
VoroNav & -- & 42.0 & -- & 0.260 \\
VLFM & 36.4 & 52.5 & 0.175 & 0.304 \\
InstructNav & -- & 58.0 & -- & 0.209 \\
UniGoal & 41.0 & 54.5 & 0.164 & 0.251 \\
\textbf{SSR-ZSON (Ours)} & \textbf{59.5} & \textbf{65.7} & \textbf{0.345} & \textbf{0.391} \\
\bottomrule
\end{tabular}
\end{table}

\subsection{Real-World Experiment}
Our real physical platform is equipped with a 12-core Arm Cortex-A78AE v8.2 64-bit processor, 64GB of RAM, 275TOPS of AI computing power, and Ubuntu 20.04 LTS. An Intel RealSense D435i depth camera and a Livox-Mid360 lidar sensor were installed on the platform.

\begin{figure}[htbp]
\centering
\includegraphics[width=\linewidth]{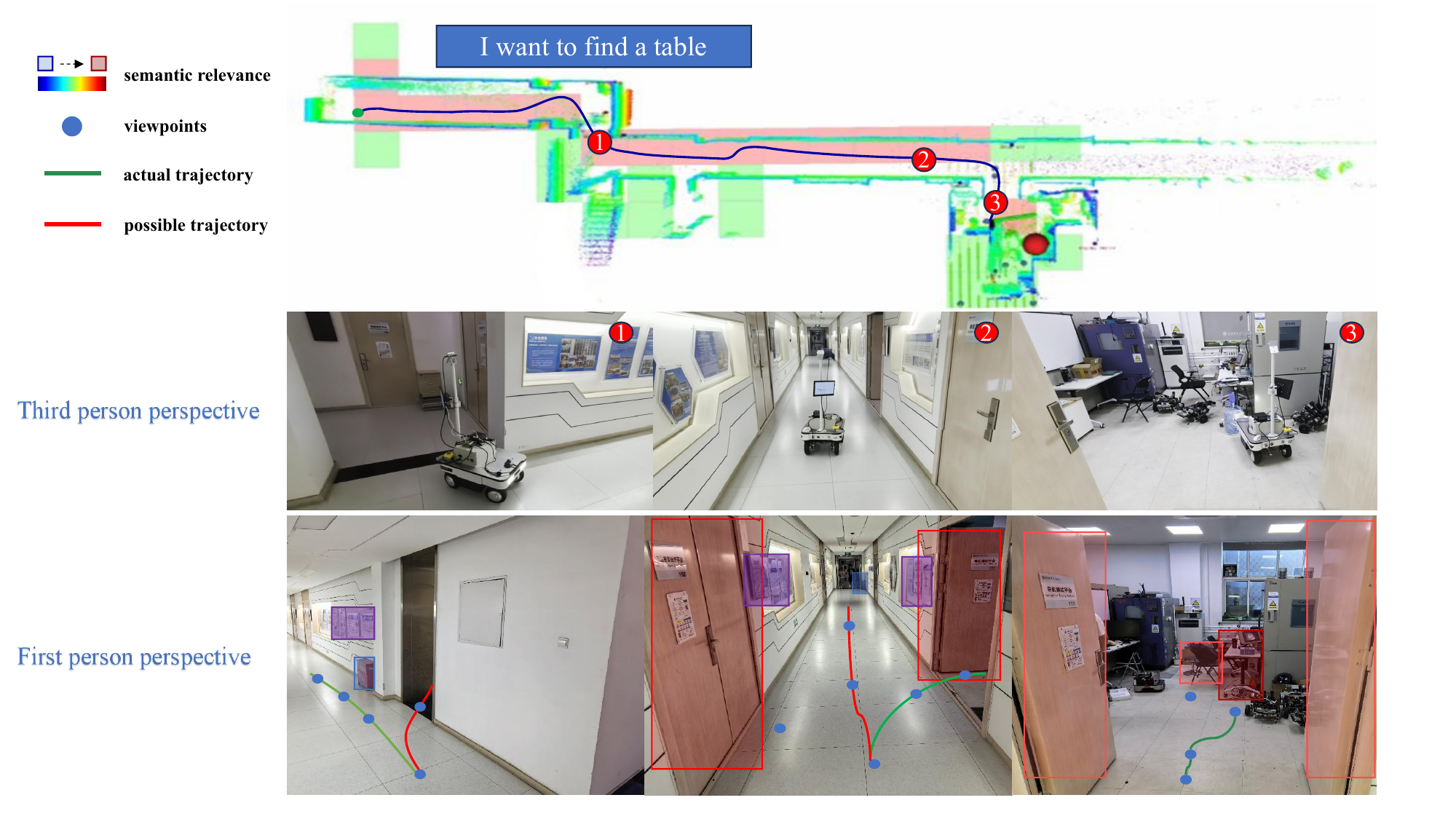}
\caption{\textbf{the process and outcomes of the physical
verification.} The experimental platform is physically validated in a long corridor containing multiple fork roads and rooms.}
\label{figure7}
\end{figure}

The process and outcomes of the physical verification conducted on the experimental platform within a long corridor featuring multiple branches and rooms are illustrated in \cref{figure7}. Upon receiving natural language instructions—such as the task of locating a desk—the platform demonstrates a preference for navigating toward areas that contain doors, entrance signs, and office-related objects. During the experiment, regions with closed doors or obstructed entrances are not designated as exploration zones, such as the goal of locating a desk—the platform demonstrates a preference for navigating toward areas that contain doors, entrance signs, and office-related objects. During the experiment, regions with closed doors or obstructed entrances are not designated as exploration zones, and target search operations are not initiated in these areas. This approach enhances both the computational efficiency of the platform and the overall efficiency of object navigation.

\section{CONCLUSIONS}

We presented the SSR-ZSON, a hierarchical framework for zero-shot object navigation that integrates LLMs with spatial-semantic reasoning to address exploration in unknown open-world. The core innovation lies in its viewpoint-region joint assessment mechanism, which combines spatial coverage and semantic density at the viewpoint level, and LLM-derived target relevance weights at the region level, to prioritize exploration toward semantically significant areas. Additionally, a collaborative memory model reduces redundant exploration by tracking spatial coverage integrity and semantic consistency. Experimental results demonstrate the effectiveness of SSR-ZSON. On the MP3D and HM3D datasets, SSR-ZSON achieves success rates of 59.5\% and 65.7\%, respectively, while the corresponding SPL are 0.345 and 0.391 that outperform existing state-of-the-art methods. These results validate that SSR-ZSON significantly enhances both navigation success and efficiency, particularly in semantically sparse environments. The framework contributes to embodied AI by deeply combining LLM-driven semantic comprehension with robust spatial analysis, enabling real-time operation in combined simulation (Habitat-Gazebo) and physical platforms. However, the current approach relies on predefined semantic associations and static environmental representations, which may limit adaptability in highly dynamic settings. Future work will focus on extending SSR-ZSON's scene reasoning capabilities, incorporating a scene prediction module, and further improving the semantic association model's adaptability to dynamic environmental variations.

\addtolength{\textheight}{-0cm}   






\end{document}